\documentclass[preprint,review,11pt,3p]{elsarticle}

\usepackage{hyperref}
\usepackage{epsfig}
\usepackage{epstopdf}
\AppendGraphicsExtensions{.tiff}

\usepackage{graphicx}
\usepackage[cmex10]{amsmath}
\usepackage{amssymb}
\usepackage{multirow}
\usepackage{rotating}
\usepackage[table]{xcolor}
\usepackage{threeparttable}
\usepackage{url}

\usepackage{color,soul}
\sethlcolor{green}
\usepackage{algorithmic}
\usepackage{algorithm}
\usepackage{amsthm}

\theoremstyle{definition}

\newcommand{\argmax}[1]{\underset{#1}{\operatorname{arg}\,\operatorname{max}}\;}
\usepackage{verbatim} 

\begin{document}

\begin{frontmatter}



\title{Image Embedded Segmentation: Uniting Supervised and Unsupervised Objectives for Segmenting Histopathological Images}

\author[label1]{Can~Taylan~Sari}
\ead{can.sari@bilkent.edu.tr}
\author[label2]{Cenk~Sokmensuer}
\ead{csokmens@hacettepe.edu.tr}
\author[label1,label3]{Cigdem~Gunduz-Demir\corref{cor1}}
\ead{gunduz@cs.bilkent.edu.tr}
\address[label1]{Department of Computer Engineering, Bilkent University, Ankara TR-06800, Turkey}
\address[label2]{Department of Pathology, Hacettepe University, Ankara TR-06100, Turkey}
\address[label3]{Neuroscience Graduate Program, Bilkent University, Ankara TR-06800, Turkey}
\cortext[cor1]{Corresponding author}

\begin{abstract}
This paper presents a new regularization method to train a fully convolutional network for semantic tissue segmentation in histopathological images. This method relies on the benefit of unsupervised learning, in the form of image reconstruction, for network training. To this end, it puts forward an idea of defining a new embedding that allows uniting the main supervised task of semantic segmentation and an auxiliary unsupervised task of image reconstruction into a single one and proposes to learn this united task by a single generative model. This embedding generates an output image by superimposing an input image on its segmentation map. Then, the method learns to translate the input image to this embedded output image using a conditional generative adversarial network, which is known as quite effective for image-to-image translations. This proposal is different than the existing approach that uses image reconstruction for the same regularization purpose. The existing approach considers segmentation and image reconstruction as two separate tasks in a multi-task network, defines their losses independently, and combines them in a joint loss function. However, the definition of such a function requires externally determining right contributions of the supervised and unsupervised losses that yield balanced learning between the segmentation and image reconstruction tasks. The proposed approach provides an easier solution to this problem by uniting these two tasks into a single one, which intrinsically combines their losses. We test our approach on three datasets of histopathological images. Our experiments demonstrate that it leads to better segmentation results in these datasets, compared to its counterparts.
\end{abstract}




\end{frontmatter}


\section{Introduction}
\label{sec:Introduction}

Unsupervised learning has been used as a regularization tool to train a neural network for a supervised task. Earlier studies have used layer-wise unsupervised pretraining to initialize weights, which are then finetuned by supervised training using backpropagation. This pretraining may provide regularization on backpropagation by enabling it to start with a better solution, and may improve the network's generalization ability~\cite{erhan10}. On the other hand, it has been argued that the weights learned by pretraining may be easily overwritten during supervised training~\cite{zhao15} or even they may not provide a better initial solution at all~\cite{rasmus15}, since the network is pretrained independently and by being unaware of the supervised task.

For more effective regularization, recent studies have trained a multi-task network to simultaneously minimize supervised and unsupervised losses by backpropagation~\cite{zhao15,zhang16}. They define the supervised loss on the main classification task and the unsupervised loss on an auxiliary image reconstruction task. These two tasks typically share an encoder path to extract feature maps, from which a decoder path reconstructs an image and a classification path estimates a one-hot class label. In~\cite{robert18}, in addition to this, another autoencoder with its own encoder and decoder is used and the outputs of the two decoders are combined to reconstruct the image. These studies calculate the reconstruction loss between original and decoded images as well as between the maps of the corresponding intermediate layers of the encoder and decoder. In~\cite{rasmus15}, noisy original images are used as inputs and the reconstruction loss is calculated between these images and their denoised versions. 

All these studies define losses on the classification and reconstruction tasks separately and linearly combine them in a joint loss function, which they use to simultaneously learn these two tasks. This may provide regularization since the tasks compete during backpropagation. On the other hand, the effectiveness of this regularization highly depends on to what extent the supervised and unsupervised losses contribute to the joint loss function. When the unsupervised loss contributes too much, the network may not sufficiently learn the main classification task. When it contributes too small, the network may not learn the auxiliary reconstruction task, which results in not getting the expected regularization effect from unsupervised learning. Thus, these studies necessitate externally selecting right contributions that yield balanced learning between the supervised and unsupervised tasks. However, depending on the application, this external selection may not be always straightforward. It may become even harder when the joint loss includes more than one reconstruction loss (e.g., the one at the input level and those at the intermediate layers).

In response to these issues, this paper introduces an easier but more effective solution to combine the supervised and unsupervised losses to train a fully convolutional network for the task of semantic segmentation in histopathological images. This solution relies on defining a new embedding that unites the main task of segmentation and an auxiliary task of image reconstruction into a single task and learning this united task by a single generative model. To this end, it first introduces an embedding that generates a multi-channel output image, on which segmentation is trivial, by superimposing an input image on its segmentation map. Then, it proposes to learn this newly generated output image from the input image using a conditional generative adversarial network (cGAN), which is known to be effective for image-to-image translations. 

This new embedding together with its learning by a cGAN provide two main advantages. First, the proposed embedding unites segmentation and reconstruction tasks, which concomitantly results in combining supervised and unsupervised objectives (losses) in a very natural way. This presents an alternative to externally determining the contributions of these tasks in a joint loss function. More importantly, since the output image of the united task corresponds to a segmentation map that preserves a reconstructive ability, uniting the segmentation and reconstruction tasks enforces the network to jointly learn image features and context features. This joint learning provides effective regularization. Second, the proposed method learns the output image of the united task by benefiting from the well-known synthesizing ability of cGANs. Thanks to using a cGAN, the method produces more realistic outputs that adhere to spatial contiguity without any postprocessing (e.g., using conditional random fields, CRFs~\cite{chen15}). To the best of our knowledge, this is the first proposal of using a cGAN to produce such embedded output images that can be directly used for semantic segmentation. Working on three datasets of histopathological images, our experiments demonstrate that the introduction of this new embedding and the proposal of learning it with a cGAN lead to successful segmentation in histopathological images, improving the results of its counterparts.

\section{Related Work}

\underline{Fully convolutional networks (FCNs)} provide efficient solutions for semantic segmentation~\cite{long15}. They are based on constructing a decoder network, consisting of deconvolution and upsampling layers, on the top of the convolutional and max-pooling layers of the encoder network~\cite{noh15}. There have been different ways to perform the deconvolution and upsampling operations. The SegNet model performs non-linear upsampling by using the max-pooling indices of its encoder also in its decoder~\cite{badrinarayanan17}. The DeepLab model uses atrous convolutions to expand the field-of-view of filters used in any (de)convolution layers~\cite{chen18}. The UNet model defines long-skip connections between the corresponding encoder and decoder layers to provide better inputs to its deconvolution operations~\cite{ronneberger15}.  It is also possible to stack multiple decoder networks consecutively to incorporate more contextual information and obtain better representations~\cite{fu19}. To regularize the training of these single-task FCN architectures, it has been proposed to use multi-task networks that consider complementary tasks along with the main task of segmentation. These are the networks with a shared encoder and parallel decoders, one for each task, and they are trained to minimize the joint loss defined on all decoders~\cite{chen17}. Alternatively, the ensemble method presented in~\cite{peng20} regularizes the FCN training by co-training two separate networks using two independent subsets of an annotated dataset.

Another way of regularization is to use unsupervised learning in the form of defining an additional image reconstruction task and learning it concurrently with the main task. In~\cite{zhao15,rasmus15}, the networks are trained to simultaneously minimize the supervised loss defined on image classification and the unsupervised loss defined on reconstructing an input image. In~\cite{zhang16}, in addition to these losses, the reconstruction errors between the maps of the corresponding encoder and decoder layers are taken into account. These previous studies focus on non-dense prediction, defining their main task as to predict one-hot class label for an entire image.  Only a few studies consider the main task of image segmentation~\cite{sun19,nguyen19}. However, all these studies use image reconstruction as an auxiliary task and linearly combine its loss and the loss of classification/segmentation, which are defined independently, in a joint loss function. This is different than our approach, which unites the image reconstruction and segmentation tasks through its proposed embedding and trains its network to minimize the loss on this united task. Moreover, these previous studies do not use a GAN for their network. 

FCNs are typically trained to predict pixel labels independent of each other. This may prevent to capture local and global spatial contiguity within an entire image. To recover fine details, CRFs using pair-wise potentials have been employed as a post-processing step to refine the segmentation maps generated by FCNs~\cite{chen15,noh15}. Although CRFs lead to improvements, the integration of FCNs and CRFs with higher orders is limited~\cite{arnab16}. This limitation has led researchers to use GANs for this purpose~\cite{luc16}.

\underline{Generative adversarial networks} are firstly proposed for image synthesis by using two networks, generator and discriminator, trained in an adversarial manner. Its application to semantic segmentation typically provides an additional input to the generator (segmentor) to control its output~\cite{luc16,isola17}. Adversarial loss has also been used to regularize network training. One work~\cite{makhzani15} uses it for an autoencoder to better learn its feature maps. It considers the encoder as the generator and feeds its outputs to the discriminator. Then, it updates encoder weights considering the adversarial loss in addition to the reconstruction loss between encoder's input and decoder's output. Another work~\cite{zhu19} estimates a segmentation map from an image and then reconstructs the image from the estimated map for regularization. It uses a cGAN for image reconstruction, and hence, employs the adversarial loss in addition to the segmentation and image reconstruction losses. However, it also separately defines these losses and linearly combines them in a joint loss function. In~\cite{yang17}, it has been proposed to concatenate the outputs of different branches, each of which inputs the upscaled map of a different decoder's intermediate layer, for liver segmentation in 3D CT volumes. This work also linearly combines the losses of these branches and uses adversarial training to better learn the final segmentation. However, none of these previous studies exploit an embedding to combine supervised and unsupervised tasks (losses) for regularizing their network for semantic segmentation and uses a cGAN for better learning the united task.

\underline{Histopathological image segmentation} has been studied at different levels. At the tissue level, the aim is to divide an image into histologically meaningful tissue compartments. Earlier studies train a CNN on image patches and then classify an image with either this CNN~\cite{xu16a} or another classifier trained on its feature maps~\cite{xu17}. Since a CNN predicts a single label for the entire image, a sliding window is usually used to label image pixels. Pixel-level predictions are also inferred using another network trained on the CNN's posteriors~\cite{chan19} and its feature maps~\cite{takahama19}. Recent studies construct a UNet model to predict pixel labels~\cite{debel18,oskal19}. It has been also proposed to fuse the predictions of multiple FCNs. In~\cite{wang16}, FCNs are trained on images of different resolutions. In~\cite{phillips19}, they are constructed by starting the upsampling operation from different layers of the same encoder. Other studies perform segmentation at finer-levels; they usually segment nucleus and gland instances. They typically use multi-task networks, in which auxiliary tasks are defined as predicting boundary of instances~\cite{chen17} and their bounding boxes~\cite{xu16b}. Application specific additional tasks, such as lumen prediction~\cite{graham19} and malignancy classification~\cite{bentaieb16}, are also used for gland instance segmentation. \textit{Note that the focus of our paper is compartment segmentation at the tissue level but not instance segmentation.}

Different than our proposal, none of these studies define an embedding to unite the segmentation and image reconstruction tasks and use a cGAN to learn this united task. Only a few use a cGAN for nucleus and gland segmentation~\cite{mahmood19,mei19}. However, these studies define adversarial loss on the genuineness of their segmentation maps but they do not consider image reconstruction loss in their segmentation networks. Besides, they do not use any embedding to regularize training. GANs are also used to synthesize additional training data~\cite{bi18}.

\begin{figure*}
\centering
\includegraphics[width=\columnwidth]{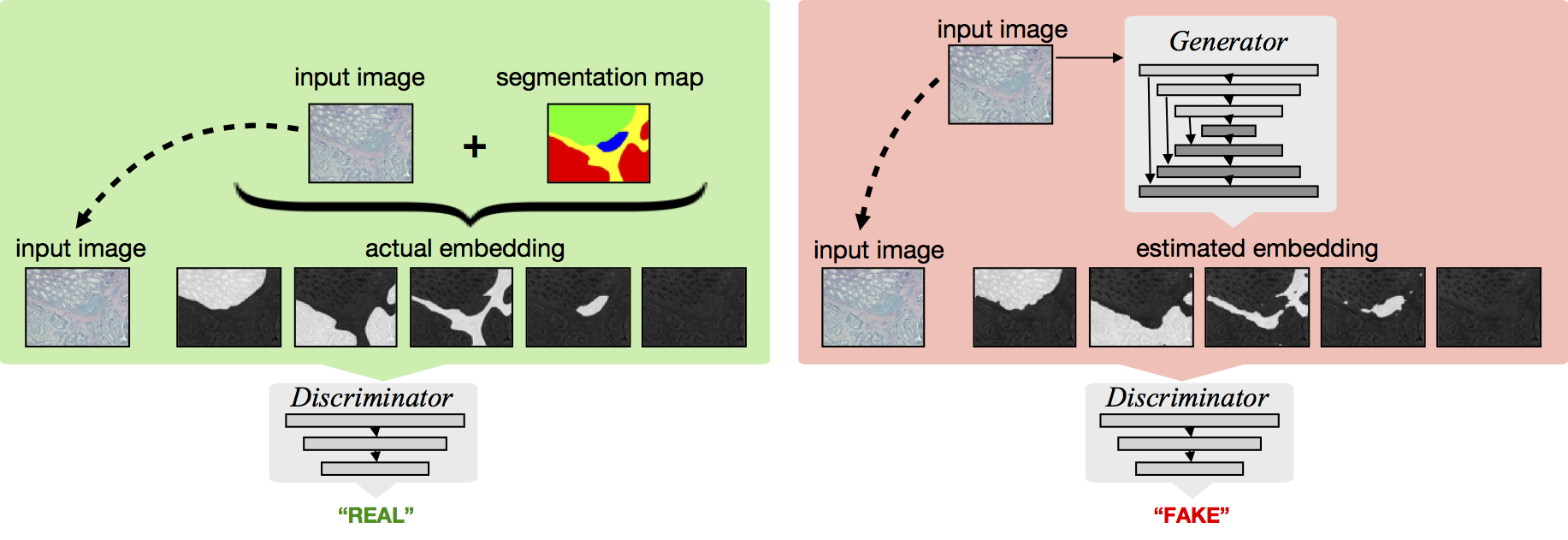}\vspace{-0.10cm}
\caption{Schematic overview of the training phase. It generates a multi-channel output image for each training instance by embedding an input image onto its segmentation map. Each channel corresponds to a segmentation label. Original input images and their generated outputs are fed to the cGAN for its training.}
\label{fig:overview}
\end{figure*}

\section{Methodology}
The proposed method, which we call \textit{the \textbf{iM}age \textbf{EM}bedded \textbf{S}egmentation (iMEMS) method}, defines a new embedding to transform semantic segmentation to the problem of image-to-image translation and solves it using a cGAN. Its motivation is as follows: The proposed transformation facilitates an easy and effective way of uniting a supervised task of semantic segmentation and an unsupervised task of image reconstruction into a single task. By its definition, learning this united task inherently requires meeting the supervised and unsupervised objectives simultaneously. Thus, the network should jointly learn image features to segment an image and context features to reconstruct it. This joint learning stands as an effective means of regularizing the network training.

The training phase starts with generating a multi-channel output image for each training instance. Then, original input images together with their generated outputs are fed to the cGAN for its training (Fig.~\ref{fig:overview}). Afterwards, the output of an unsegmented image is estimated by the generator of the trained cGAN. The details are given in the following sections. The iMEMS method is implemented in Python using the Keras framework. The source codes are available at http://www.cs.bilkent.edu.tr/$\sim$gunduz/downloads/iMEMS.

\begin{figure*}
\centering
\small{
\begin{tabular}{@{~}c@{~}c@{~}c@{~}c@{~}c@{~}}
\includegraphics[width=0.19\columnwidth]{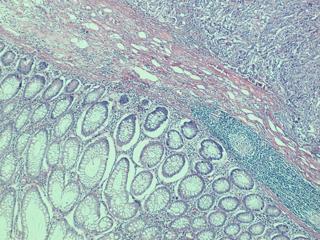} &
\includegraphics[width=0.19\columnwidth]{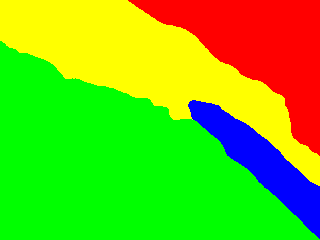} &
\includegraphics[width=0.19\columnwidth]{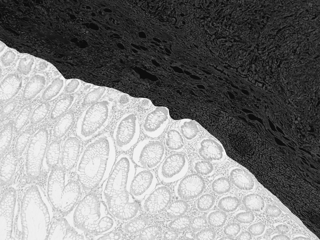} &
\includegraphics[width=0.19\columnwidth]{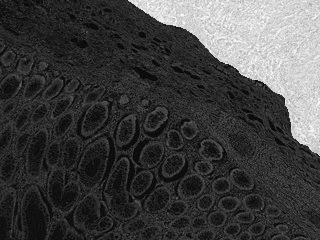} &
\includegraphics[width=0.19\columnwidth]{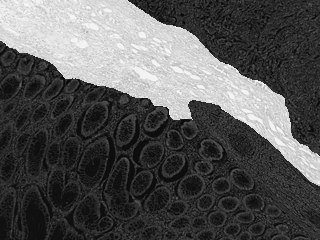} \\
(a) & (b) & (c) & (d) & (e)
\end{tabular}
}
\caption{(a) An original input image $I$. (b) Its ground truth segmentation map $S_I$. (c) The first, (d) second,  and (e) third channels in its output image, which are generated for the segmentation label shown as green, red, and yellow in $S_I$, respectively. Note that this semantic segmentation problem is a task of predicting one of the five labels for each pixel; this particular image does not contain any pixel belonging to the fifth label. Thus, the generated output image $O_I$ has five channels (i.e., $O_{I}^{[1]}$, $O_{I}^{[2]}$, $O_{I}^{[3]}$, $O_{I}^{[4]}$, and $O_{I}^{[5]}$ are generated for the input image). This figure shows only three of these channels.}
\label{fig:embedding-examples}
\end{figure*}

\subsection{Proposed Embedding}
\label{sec:embedding} 

Let $I$ be an RGB image in the training set, $G_I$ be its grayscale, and $S_I$ be its ground truth segmentation map that may contain $K$ possible labels. This embedding generates a $K$-channel output image $O_I$ by superimposing the grayscale $G_I$ on the segmentation map $S_I$. For that, for each segmentation label $k \in \{1, ..., K\}$, it generates an output channel $O_{I}^{[k]}$. For a pixel $p$, this output channel is defined as follows:
\begin{equation}
\label{eqn:embedding}
O_{I}^{[k]}(p)=
\begin{cases}
	\left\lfloor\frac{G_I(p)}{2}\right\rfloor + 128 	& \text{~~~~~~if}\ S_I(p) = k \vspace{0.3cm}\\    
	127 - \left\lfloor\frac{G_I(p)}{2}\right\rfloor 	& \text{~~~~~~if}\ S_I(p) \neq k \vspace{0.3cm}\\    
   \end{cases}
\end{equation}
This definition maps grayscale intensities of all pixels belonging to the $k$-th label to the interval of [128, 255] in the $k$-th output channel $O_{I}^{[k]}$ and to the interval of [0, 127] in all other channels. However, in mapping these intensities to [0, 127], it inverts their values to make the characteristics of pixels in foreground and background regions of the $k$-th channel more distinguishable. In other words, a grayscale intensity interval [0, 255] is mapped to [128, 255] in the $k$-th output channel if a pixel belongs to the $k$-th label, and to [127, 0] otherwise. Note that this definition equally divides the grayscale interval to represent pixels in foreground and background regions in the $k$-th channel. This is an appropriate choice for our application since each channel needs to represent two types of regions (i.e., background and foreground regions). However, this definition can easily be modified such that it uses unequal divisions of the interval, if this is necessary for other applications. 

This definition is illustrated in Fig.~\ref{fig:embedding-examples}. As seen here, foreground regions in each channel seem brighter, as they are mapped to [128, 255], whereas background regions seem darker, as they are mapped to [0, 127]. Thus, it is trivial to segment foreground regions in each channel of this output image. Besides, both foreground and background regions in this output preserve the original image content, which helps regularize a network in learning how to distinguish these two regions.

\subsection{cGAN Architecture and Training}
\label{sec:cgan}
The definition in Eqn.~\ref{eqn:embedding} requires the ground truth map $S_I$ for an input image $I$. Thus, the iMEMS method only employs this definition to generate the output images for segmented training instances, which are used to train a cGAN. Then, for an unsegmented (test) image, iMEMS estimates this output from an original input image using the trained cGAN. In other words, it translates one image to another using a cGAN.

\begin{figure*}
\centering
\includegraphics[width=\columnwidth]{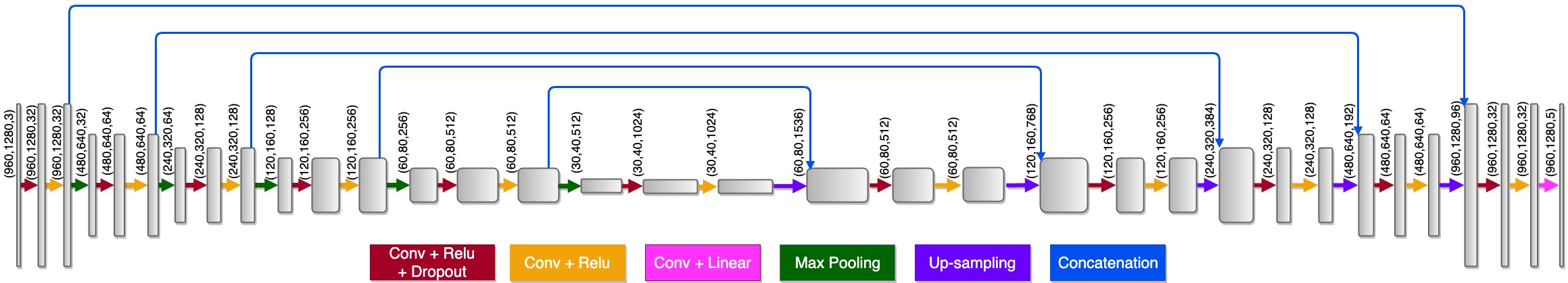}
\vspace{-0.5cm}
\caption{Architecture of the generator network in the cGAN. Different layers and operations are indicated with different colors. The resolution of the feature maps in each layer together with the number of these feature maps are also indicated.}
\label{fig:generator}
\end{figure*}

The generator of this cGAN inputs a normalized RGB image $I$ and outputs a $K$-channel image $\widehat{O}_I$. It uses a UNet architecture with an encoder and a decoder connected by symmetric connections (Fig.~\ref{fig:generator}). The convolution layers, except the last one, use $3\times3$ filters and the ReLU activation function. The last layer uses a linear function since it estimates continuous intensity values of the output image. The pooling/upsampling layers use $2\times2$ filters. Extra dropout layers are added to reduce overfitting; the dropout factor is set to 0.2.

\begin{figure}
\centering
\includegraphics[width=0.6\columnwidth]{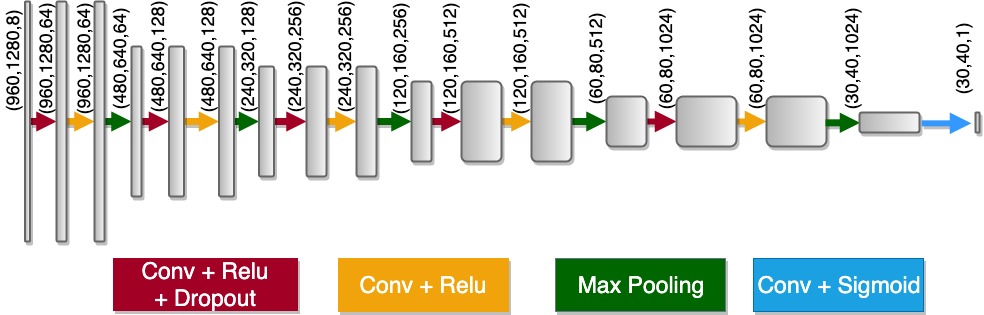}
\caption{Architecture of the discriminator network in the cGAN.}
\label{fig:discriminator}
\end{figure}

The discriminator inputs a normalized RGB image and the $K$-channel output image corresponding to this input. Its output is a class label to indicate whether the output image is real or fake; i.e., it estimates if this output is calculated by Eqn.~\ref{eqn:embedding} using the ground truth or produced by the generator. Its architecture is given in Fig.~\ref{fig:discriminator}. It has the same operations with the generator's encoder except that its last layer uses the sigmoid function. This network uses a convolutional PatchGAN classifier~\cite{isola17}, which uses local patches to determine whether the output image is real or fake rather than the entire image. 


The generator and discriminator networks are trained from scratch. The batch size is 1. The network weights are learned on the training images for 300 epochs. At each epoch, the loss is calculated on the validation images and the network that gives the minimum validation loss is selected at the end. The loss settings of this cGAN are the same with~\cite{isola17}. The objective function is $\mathrm{arg} \mathrm{min}_{G} \mathrm{max}_{D}~\mathcal{L}_{adv}(G,D) + \lambda~\mathcal{L}_{L1}(G)$, where $\mathcal{L}_{adv}(G,D)$ is the adversarial loss on the discriminator's outputs and $\mathcal{L}_{L1}(G)$ is the L1 loss on the generator's output. Similar to~\cite{isola17}, the weight $\lambda$ of the L1 loss is selected as 100. It is worth to noting that although this objective linearly combines two losses, its purpose is different than the proposed iMEMS method. As opposed to iMEMS, this objective does not directly aim to combine the losses of the supervised task of semantic segmentation and the unsupervised task of image reconstruction. The iMEMS method defines an embedding to unite these two tasks into a single one and uses a cGAN for better learning this united task. Indeed, both the generator and the discriminator of the cGAN define their tasks on the united task of the iMEMS method, which means the adversarial and L1 losses are also defined on this united task.

\subsection{Tissue Segmentation}
\label{sec:classify} 

\begin{figure*}
\centering
\small{
\begin{tabular}{@{~}c@{~}c@{~}c@{~}c@{~}c@{~}}
\includegraphics[width=0.19\columnwidth]{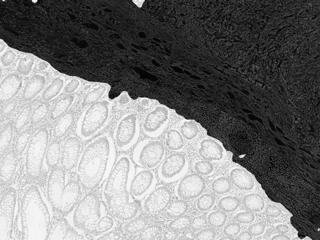} &
\includegraphics[width=0.19\columnwidth]{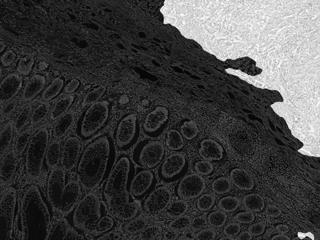} & 
\includegraphics[width=0.19\columnwidth]{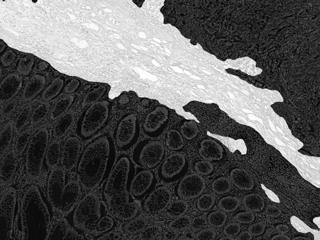} &
\includegraphics[width=0.19\columnwidth]{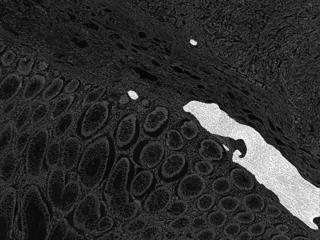} &
\includegraphics[width=0.19\columnwidth]{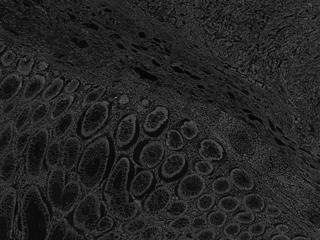}
\vspace{-0.5cm}
\end{tabular}
}
\caption{Output maps $\widehat{O}_U^{[k]}$ estimated by the generator of the cGAN for the image shown in Fig.~\ref{fig:embedding-examples}.}
\label{fig:example}
\end{figure*}

For an unsegmented image $U$, the iMEMS method estimates the output $\widehat{O}_U$ using the generator of the trained cGAN and segments it based on this estimated output. In particular, it classifies each pixel $p$ with a segmentation label $k$ whose corresponding output has the highest estimated value; that is, $\widehat{S}_U(p) = \argmax{k} \widehat{O}_U^{[k]}(p)$. For the image shown in Fig.~\ref{fig:embedding-examples}, the estimated output images are illustrated in Fig.~\ref{fig:example}.

\section{Datasets}

We test the iMEMS method on three datasets that contain microscopic images of hematoxylin-and-eosin stained tissues. The first one is an in-house colon dataset and the other two are publicly available epithelium and tubule datasets, which are prepared by another research group~\cite{janowczyk16}. 

The \textbf{\textit{in-house dataset}} contains 365 images of colon tissues collected from the Pathology Department Archives of Hacettepe University. Images are scanned at $5\times$, using a Nikon Coolscope Digital Microscope. Image resolution is $960\times1280$. In each image, non-overlapping regions are annotated by C. Sokmensuer, who is an experienced board-certified pathologist, with one of the five labels: normal, tumorous (colon adenocarcinomatous), connective tissue, dense lymphoid tissue, and non-tissue (empty glass and debris). In this dataset, 100 images are randomly selected as training instances. The remaining ones are used as test instances, on which we measure the performance of our method and comparison algorithms. This dataset is available upon request.

The \textbf{\textit{epithelium dataset}} consists of 42 estrogen receptor positive breast cancer images scanned at $20\times$. Image resolution is $1000\times1000$. In each image, non-overlapping regions are annotated as either epithelium or background~\cite{janowczyk16}. Since the size of this dataset is relatively small, we randomly split it into five folds and measure the performance using five-fold cross-validation. Furthermore, we divide an image of each fold into four equal non-overlapping parts in order to make images optimal for the proposed architecture and also to increase the number of training instances. Note that all four parts belonging to the same image are used in the same fold. 

The \textbf{\textit{tubule dataset}} consists of 85 colorectal images scanned at $40\times$. As these images have different resolutions, we rescale them to $522\times775$ pixels, which is the resolution of more than 90 percent of all images. In each image, tubule and background regions are annotated~\cite{janowczyk16}. Likewise, the size of this dataset is also relatively small. Thus, we also use five-fold cross-validation to assess the methods' performance.

\section{Results}

Two metrics are used for quantitative evaluation. The first one is the pixel-level accuracy, which gives the percentage of correctly predicted pixels in all images. The second one is the pixel-level F-score that is calculated for each segmentation label separately. That is, for each label, the F-score is calculated considering the pixels of this label as positive and those of the other label(s) as negative. The average of these class-wise F-scores is also calculated. The quantitative results are reported in Table~\ref{table:results}. In this table, the metrics are calculated on the test set images for the in-house colon dataset. For the other two datasets, these are the average test fold metrics calculated over five runs (using five-fold cross-validation). Note that, for each run, the method of interest is trained on the images of four out of five folds and the remaining one is considered as the test fold. These results show that the proposed iMEMS method gives high F-scores for all segmentation labels, leading to the best accuracy and the best average F-score, for all datasets. 
\begin{table}[t!]
\caption{F-scores and accuracies of the proposed iMEMS method and the comparison algorithms. (a) For the in-house colon dataset, these metrics are obtained on the test set. (b) For the epithelium dataset, these are the average metrics obtained on the five test folds. (c) For the tubule dataset, these are also the average metrics obtained on the five test folds.}
\centering
\small{
\begin{tabular}{c}
\begin{tabular}{|l|c|c|c|c|c|c|c|}
\hline
& \multicolumn{6}{c|}{ \cellcolor[gray]{0.75}\textbf{F-scores}} & \cellcolor[gray]{0.75} \\ \cline{2-7}
& \textit{Normal}	& \textit{Tumorous}	& \textit{Connective}		& \textit{Lymphoid}	& \textit{Non-tissue}	
& \textit{Average}	& \cellcolor[gray]{0.75}\textbf{\footnotesize{Accuracy}}	\\ \hline
iMEMS				& \textbf{94.81} & \textbf{93.12} & \textbf{84.43} & \textbf{80.54} & \textbf{86.00} & \textbf{87.78}	& \textbf{91.76}	\\ \hline
\texttt{UNet-C-single}				& 92.89 		& 91.83 		& 79.96 		& 77.55 		& 61.28 		& 80.70 		& 89.27		\\ \hline
\texttt{cGAN-C-single}				& 92.45 		& 90.65 		& 76.74 		& 78.87 		& 80.33 		& 83.81 		& 88.49		\\ \hline
\texttt{UNet-R-single}				& 93.12 		& 91.17 		& 75.72 		& 72.78 		& 78.29 		& 82.22 		& 88.80		\\ \hline
\texttt{UNet-C-multi}					& 92.89 		& 91.85 		& 82.00 		& 78.91 		& 77.83 		& 84.70 		& 89.91		\\ \hline
\footnotesize{\texttt{UNet-C-multi-int}}	& 90.43 		& 89.80 		& 80.49 		& 79.13 		& 83.09 		& 84.59 		& 88.03		\\ \hline
\end{tabular} \\
(a) \vspace{0.4cm}\\ 
\begin{tabular}{|l|c|c|c|c|}
\hline
& \multicolumn{3}{c|}{\cellcolor[gray]{0.75}\textbf{F-scores}} & \multicolumn{1}{c|}{\cellcolor[gray]{0.75}} \\ \cline{2-4}
&\textit{Epithelium} &\textit{Background} & \textit{Average} & \multicolumn{1}{c|}{\cellcolor[gray]{0.75}\textbf{\footnotesize{Accuracy}}} \\ \hline
iMEMS						& \textbf{85.51}	& \textbf{92.40}	& \textbf{88.96}	& \textbf{90.17}			\\ \hline
\texttt{UNet-C-single}				& 81.86	& 89.74	& 85.80	& 86.96		\\ \hline
\texttt{cGAN-C-single}				& 81.67	& 90.14	& 85.91	& 87.26			\\ \hline
\texttt{UNet-R-single} 				& 82.59	& 91.02	& 86.81	& 88.20			\\ \hline
\texttt{UNet-C-multi}					& 81.82	& 90.65	& 86.23	& 87.72			\\ \hline
\footnotesize{\texttt{UNet-C-multi-int}}	& 81.71	& 90.57	& 86.14	& 87.60			\\ \hline
\end{tabular} \\
(b) \vspace{0.4cm}\\ 
\begin{tabular}{|l|c|c|c|c|}
\hline
& \multicolumn{3}{c|}{\cellcolor[gray]{0.75}\textbf{F-scores}} & \multicolumn{1}{c|}{\cellcolor[gray]{0.75}} \\ \cline{2-4}
&\textit{Tubule~} &\textit{Background} & \textit{Average} & \multicolumn{1}{c|}{\cellcolor[gray]{0.75}\textbf{\footnotesize{Accuracy}}} \\ \hline
iMEMS				& \textbf{87.08}	& \textbf{87.00}	& \textbf{87.04} & \textbf{87.09}			\\ \hline
\texttt{UNet-C-single}				& 84.65	& 83.57	& 84.11	& 84.26		\\ \hline
\texttt{cGAN-C-single}				& 85.01	& 84.58	& 84.79	& 84.88			\\ \hline
\texttt{UNet-R-single} 				& 84.48	& 83.83	& 84.15	& 84.37			\\ \hline
\texttt{UNet-C-multi}					& 86.06	& 84.47	& 85.27	& 85.43			\\ \hline
\footnotesize{\texttt{UNet-C-multi-int}}	& 85.67	& 85.30	& 85.49	& 85.59			\\ \hline
\end{tabular} \\
(c)
\end{tabular}
}
\label{table:results} 
\end{table}

\begin{figure}[t!]
\centering
\scriptsize{
\begin{tabular}{c@{~}c@{~}c@{~}c@{~}c@{~}c@{~}c@{~}}
\begin{sideways}~~~~~~~~Image\end{sideways} &
\includegraphics[width=0.182\columnwidth]{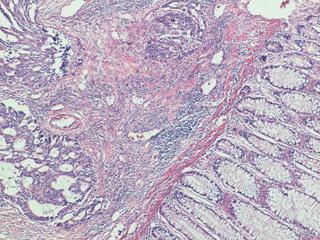} &
\includegraphics[width=0.182\columnwidth]{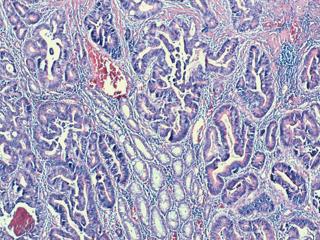} &
\includegraphics[width=0.182\columnwidth]{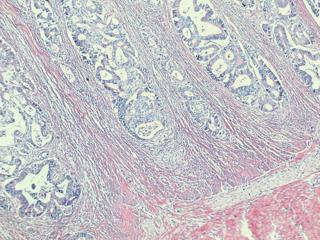} &
\includegraphics[width=0.182\columnwidth]{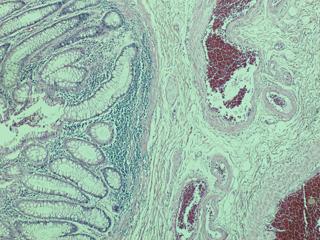} &
\includegraphics[width=0.182\columnwidth]{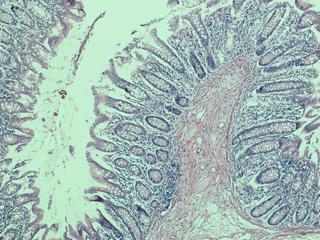} \\
\begin{sideways} ~~~~Annotation\end{sideways} &
\includegraphics[width=0.182\columnwidth]{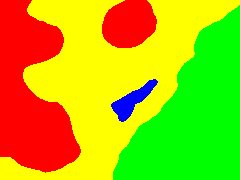} &
\includegraphics[width=0.182\columnwidth]{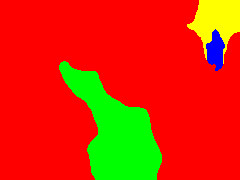} &
\includegraphics[width=0.182\columnwidth]{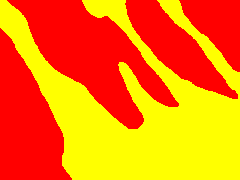} &
\includegraphics[width=0.182\columnwidth]{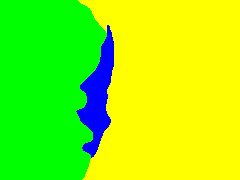} &
\includegraphics[width=0.182\columnwidth]{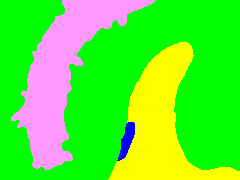} \\
\begin{sideways} ~~~~~~~iMEMS \end{sideways} &
\includegraphics[width =0.182\columnwidth]{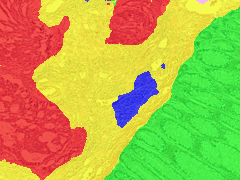} &
\includegraphics[width =0.182\columnwidth]{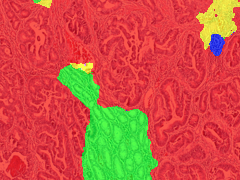} &
\includegraphics[width =0.182\columnwidth]{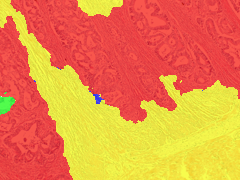} &
\includegraphics[width =0.182\columnwidth]{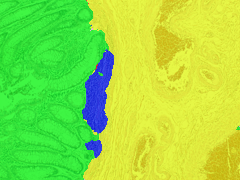} &
\includegraphics[width =0.182\columnwidth]{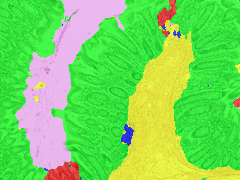} \\
\begin{sideways} ~~UNet-C-single \end{sideways} &
\includegraphics[width =0.182\columnwidth]{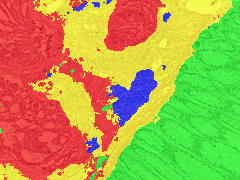} &
\includegraphics[width =0.182\columnwidth]{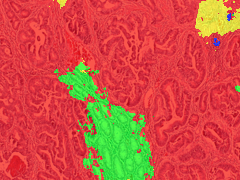} &
\includegraphics[width =0.182\columnwidth]{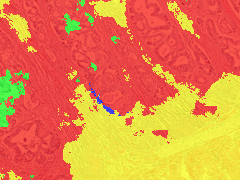} &
\includegraphics[width =0.182\columnwidth]{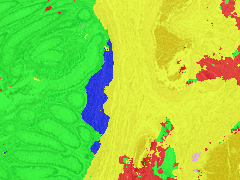} &
\includegraphics[width =0.182\columnwidth]{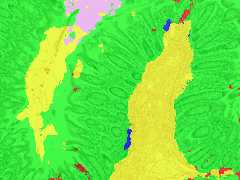} \\
\begin{sideways} ~cGAN-C-single\end{sideways} &
\includegraphics[width =0.182\columnwidth]{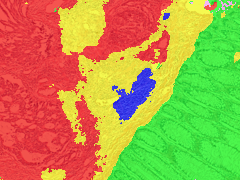} &
\includegraphics[width =0.182\columnwidth]{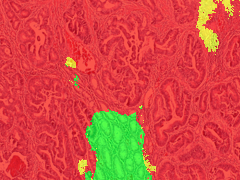} &
\includegraphics[width =0.182\columnwidth]{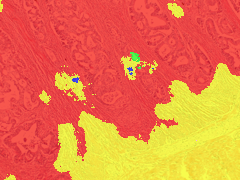} &
\includegraphics[width =0.182\columnwidth]{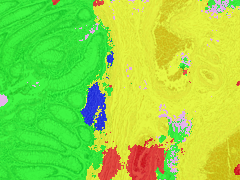} &
\includegraphics[width =0.182\columnwidth]{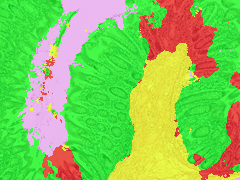} \\
\begin{sideways} ~~UNet-R-single\end{sideways} &
\includegraphics[width =0.182\columnwidth]{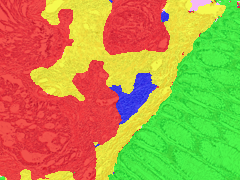} &
\includegraphics[width =0.182\columnwidth]{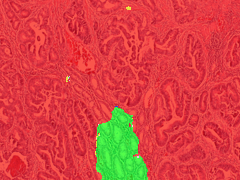} &
\includegraphics[width =0.182\columnwidth]{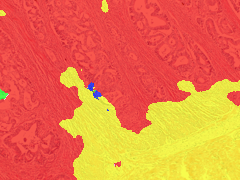} &
\includegraphics[width =0.182\columnwidth]{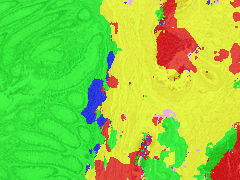} &
\includegraphics[width =0.182\columnwidth]{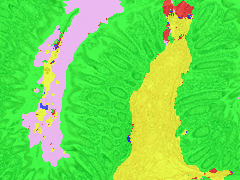} \\
\begin{sideways} ~~UNet-C-multi \end{sideways} &
\includegraphics[width =0.182\columnwidth]{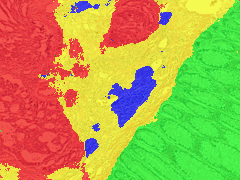} &
\includegraphics[width =0.182\columnwidth]{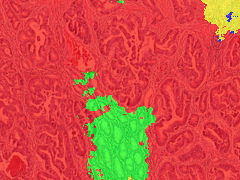} &
\includegraphics[width =0.182\columnwidth]{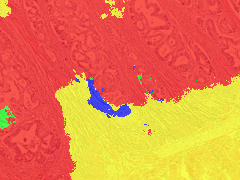} &
\includegraphics[width =0.182\columnwidth]{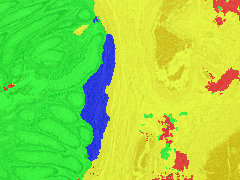} &
\includegraphics[width =0.182\columnwidth]{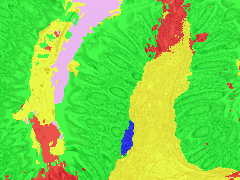} \\
\begin{sideways} UNet-C-multi-int \end{sideways} &
\includegraphics[width =0.182\columnwidth]{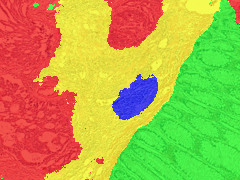} &
\includegraphics[width =0.182\columnwidth]{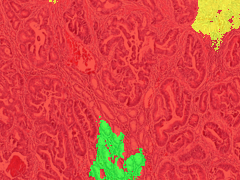} &
\includegraphics[width =0.182\columnwidth]{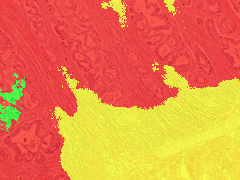} &
\includegraphics[width =0.182\columnwidth]{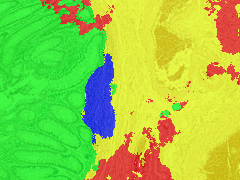} &
\includegraphics[width =0.182\columnwidth]{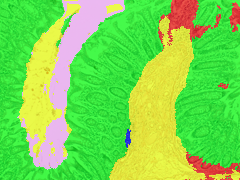}
\vspace{-0.1cm}
\end{tabular}}
\caption{For the \textit{in-house colon dataset}, visual results on example test images. Segmentation labels are shown with green (normal), red (tumorous), yellow (connective tissue), blue (dense lymphoid tissue), and pink (non-tissue). Results are embedded on original images for better visualization.}
\label{fig:results-in-house}
\end{figure}

\begin{figure}[t!]
\centering
\scriptsize{
\begin{tabular}{@{~}c@{~}c@{~}c@{~}c@{~~~~~}c@{~}c@{~}c@{~}}
\begin{sideways}~~~~~~Image\end{sideways} &
\includegraphics[height= 2.15cm]{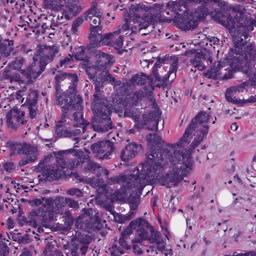} &
\includegraphics[height= 2.15cm]{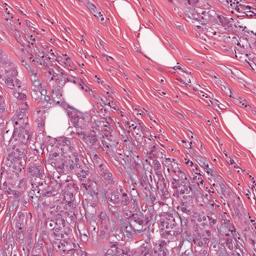} &
\includegraphics[height= 2.15cm]{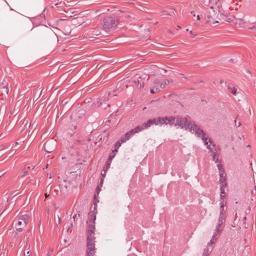} &
\includegraphics[height= 2.15cm]{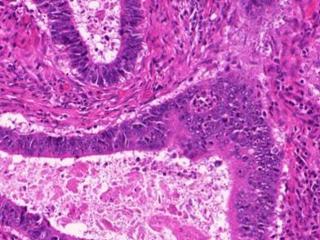} &
\includegraphics[height= 2.15cm]{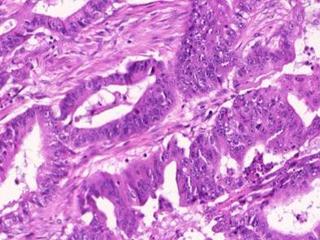} &
\includegraphics[height= 2.15cm]{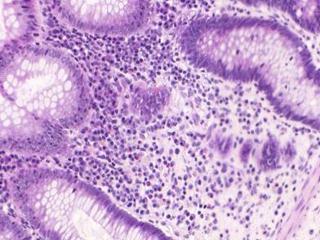} \\
\begin{sideways} ~~~Annotation\end{sideways} &
\includegraphics[height= 2.15cm]{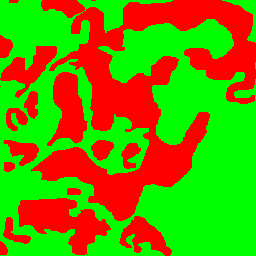} &
\includegraphics[height= 2.15cm]{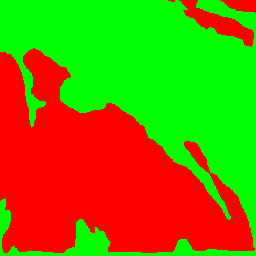} &
\includegraphics[height= 2.15cm]{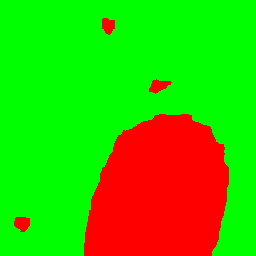} &
\includegraphics[height= 2.15cm]{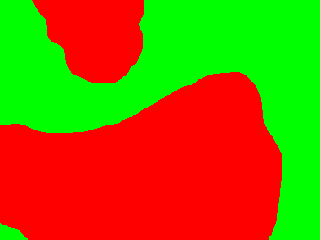} &
\includegraphics[height= 2.15cm]{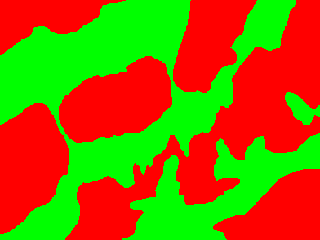} &
\includegraphics[height= 2.15cm]{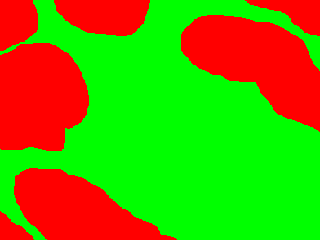} \\
\begin{sideways} ~~~~~iMEMS \end{sideways} &
\includegraphics[height= 2.15cm]{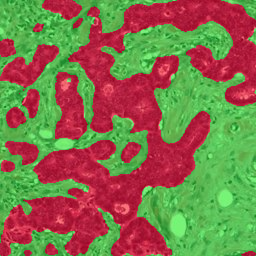} &
\includegraphics[height= 2.15cm]{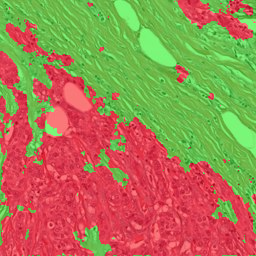} &
\includegraphics[height= 2.15cm]{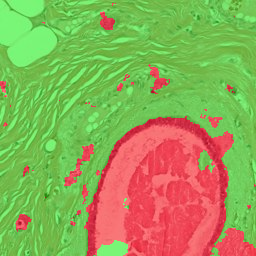} &
\includegraphics[height= 2.15cm]{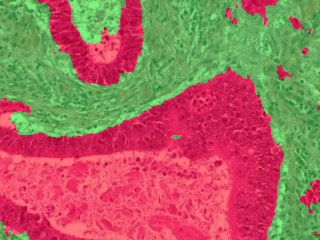} &
\includegraphics[height= 2.15cm]{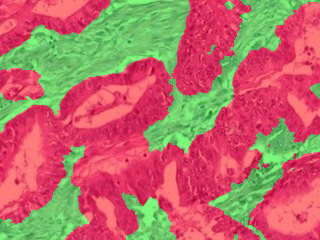} &
\includegraphics[height= 2.15cm]{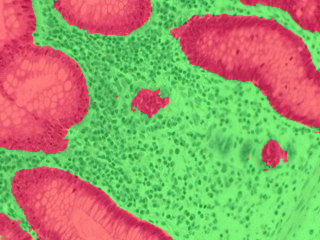} \\
\begin{sideways} ~UNet-C-single \end{sideways} &
\includegraphics[height= 2.15cm]{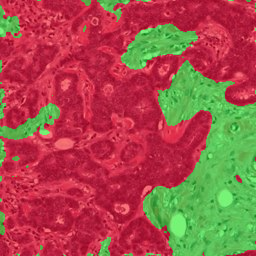} &
\includegraphics[height= 2.15cm]{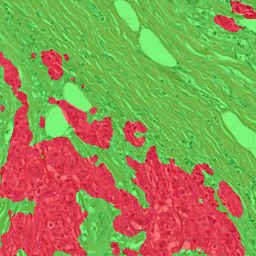} &
\includegraphics[height= 2.15cm]{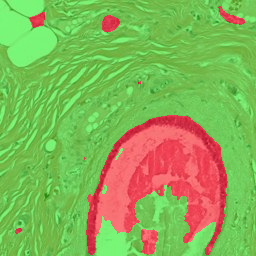} &
\includegraphics[height= 2.15cm]{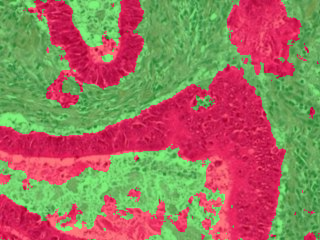} &
\includegraphics[height= 2.15cm]{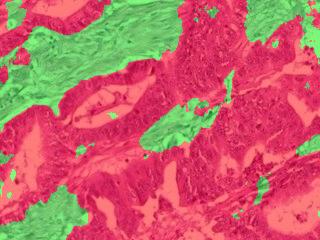} &
\includegraphics[height= 2.15cm]{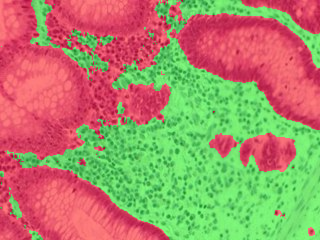} \\
\begin{sideways} ~cGAN-C-single\end{sideways} &
\includegraphics[height= 2.15cm]{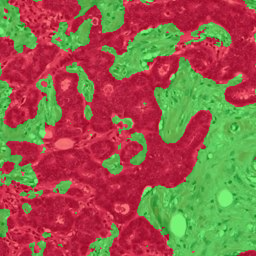} &
\includegraphics[height= 2.15cm]{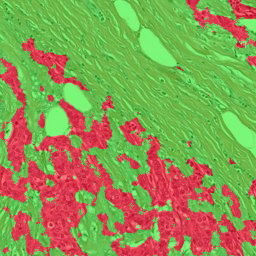} &
\includegraphics[height= 2.15cm]{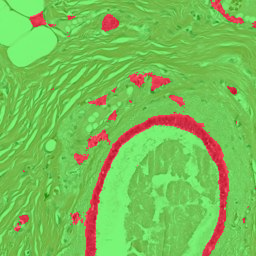} &
\includegraphics[height= 2.15cm]{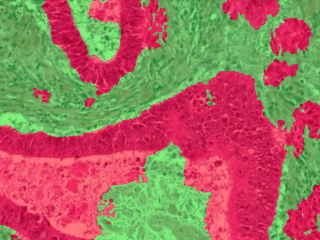} &
\includegraphics[height= 2.15cm]{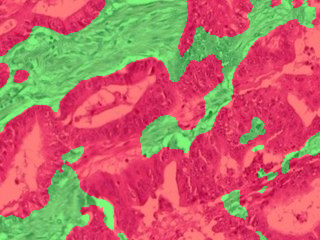} &
\includegraphics[height= 2.15cm]{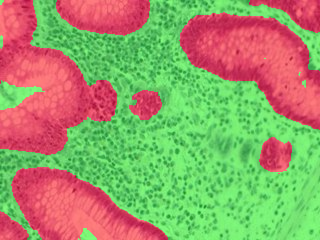} \\
\begin{sideways} ~UNet-R-single\end{sideways} &
\includegraphics[height= 2.15cm]{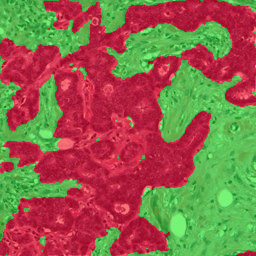} &
\includegraphics[height= 2.15cm]{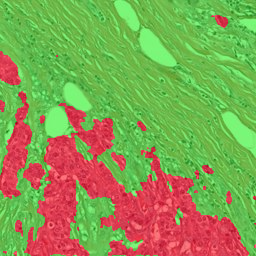} &
\includegraphics[height= 2.15cm]{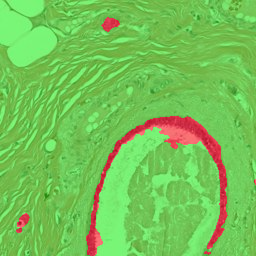} &
\includegraphics[height= 2.15cm]{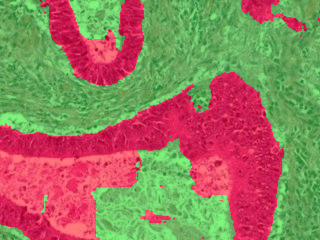} &
\includegraphics[height= 2.15cm]{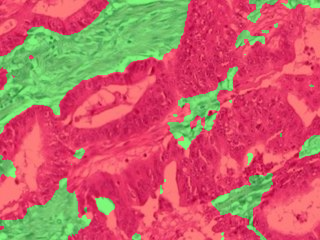} &
\includegraphics[height= 2.15cm]{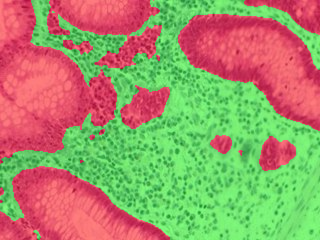} \\
\begin{sideways} ~~UNet-C-multi \end{sideways} &
\includegraphics[height= 2.15cm]{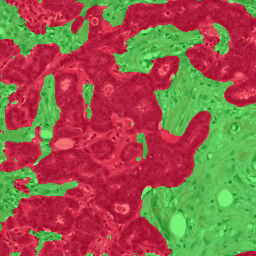} &
\includegraphics[height= 2.15cm]{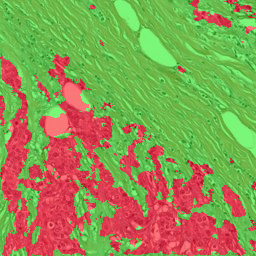} &
\includegraphics[height= 2.15cm]{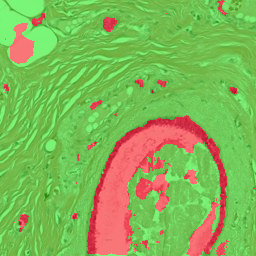} &
\includegraphics[height= 2.15cm]{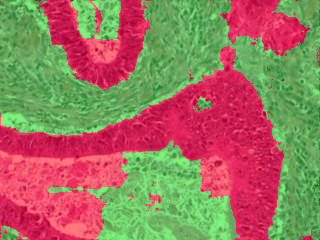} &
\includegraphics[height= 2.15cm]{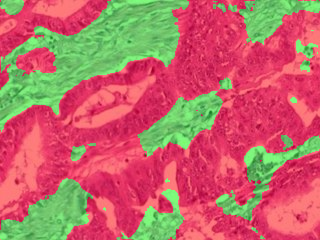} &
\includegraphics[height= 2.15cm]{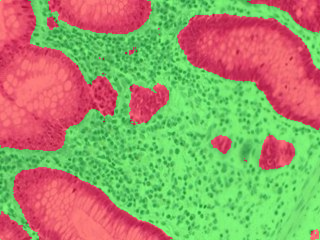} \\
\begin{sideways} UNet-C-multi-int \end{sideways} &
\includegraphics[height= 2.15cm]{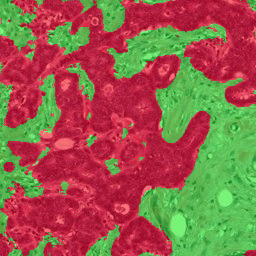} &
\includegraphics[height= 2.15cm]{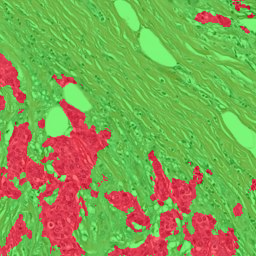} &
\includegraphics[height= 2.15cm]{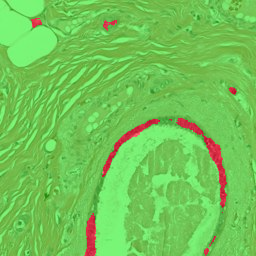} &
\includegraphics[height= 2.15cm]{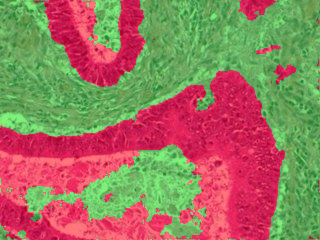} &
\includegraphics[height= 2.15cm]{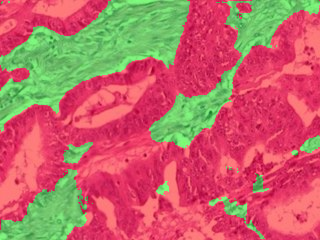} &
\includegraphics[height= 2.15cm]{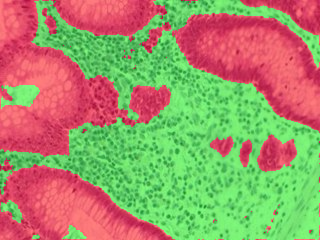}
\vspace{-0.3cm}
\end{tabular}}
\caption{\underline{First three columns}: Visual results on example test images for the \textit{epithelium dataset}. Segmentation labels are shown with red (epithelium) and green (background). \underline{Last three columns}: Visual results on example test images for the \textit{tubule dataset}. Segmentation labels are shown with red (tubule) and green (background). Results are embedded on original images for better visualization.}
\label{fig:results-others}
\end{figure}

Visual results on example test set/fold images are shown in Figs.~\ref{fig:results-in-house} and~\ref{fig:results-others}. They reveal that the iMEMS method does not only give higher performance metrics but also produces more realistic segmentations that adhere to spatial contiguity in pixel predictions, especially for the in-house colon dataset (Fig.~\ref{fig:results-in-house}). This is attributed to the effectiveness of using the proposed embedding as the output and learning it with a cGAN. Since this output also includes the original image content, it provides regularization on the segmentation task. Moreover, since the discriminator performs real/fake classification on the entire output, it enforces the generator to produce embeddings that better preserve the shapes of the segmented regions. 

To better explore these two factors (namely, using the proposed embedding and learning it with a cGAN), we compare iMEMS with five comparison algorithms summarized in Table~\ref{table:summary}. These algorithms either estimate the original segmentation map or the proposed embedding using either a UNet or a cGAN. For fair comparisons, the algorithms that use a cGAN have the same architecture with our method and those that use a UNet have the architecture of our method's generator. The last layer of a network uses a linear function if it estimates the proposed embedding, and a softmax function if it estimates the segmentation map. Two comparison algorithms use a multi-task network that concurrently learns the segmentation and image reconstruction tasks. These networks contain a shared encoder and two parallel decoders, whose architectures are the same with those of the generator. 
\begin{table*}[t!]
\caption{Summary of the algorithms used for the comparative study. The naming convention of these algorithms is x-y-z. X is the network type that the algorithm uses. Y is R (regression) if the estimated output is the proposed embedding and C (classification) if it is the segmentation map. Z indicates whether the algorithm uses a single-task or a multi-task network.}
\vspace{0.2cm}
\centering
\small{
\begin{tabular}{|l@{~}|l@{~}|l@{~}|l|}
\hline
\rowcolor[gray]{0.75}
Method name 				& Network 		& Output					& Task \\ \hline
iMEMS		 			& cGAN 			& Proposed embedding 		& Single-task regression \\ \hline
\texttt{UNet-C-single} 		& UNet 			& Segmentation map 		& Single-task classification \\ \hline
\texttt{cGAN-C-single} 		& cGAN 			& Segmentation map 		& Single task classification \\ \hline
\texttt{UNet-R-single} 		& UNet 			& Proposed embedding 		& Single-task regression \\ \hline
\multirow{2}{*}{\texttt{UNet-C-multi} } & \multirow{2}{*}{UNet} & Segmentation map and & Multi-task classification and image reconstruction \\ 
& & reconstructed image & (\textit{reconstruction loss is calculated at the input level}) \\ \hline
\multirow{3}{*}{\texttt{UNet-C-multi-int} } & \multirow{3}{*}{UNet} & Segmentation map & Multi-task classification and image reconstruction \\ 
& & reconstructed image & (\textit{reconstruction loss is calculated at the input level} \\ 
& & & \textit{as well as the intermediate layers}) \\ \hline
\end{tabular}
}
\label{table:summary} 
\end{table*}

First, we compare iMEMS with three algorithms that consider none or only one of the two factors. \texttt{UNet-C-single} is the baseline that considers none; it estimates the original segmentation map using a UNet. \texttt{cGAN-C-single} estimates the segmentation map but this time with the cGAN also used by iMEMS. \texttt{UNet-R-single} also estimates the proposed embedding but not using a cGAN. The results in Table~\ref{table:results} show that the contribution of both factors is critical to obtain the best results. Furthermore, they show that the proposed embedding provides effective regularization for network training regardless of the network type. \texttt{UNet-R-single} improves the results of \texttt{UNet-C-single} and iMEMS improves those of \texttt{cGAN-C-single}. Nevertheless, the proposed embedding together with the cGAN yields better improvement. 

Next, we compare iMEMS with another regularization technique that simultaneously minimizes supervised and unsupervised losses defined on the segmentation and image reconstruction tasks, respectively. This technique relies on constructing a multi-task network whose weights are learned by minimizing a joint loss function~\cite{zhao15,zhang16}. For the supervised loss, $\mathcal{L}_{seg}$, the average cross-entropy is used. For the unsupervised loss, two definitions are used. First is the reconstruction loss, $\mathcal{L}_{rec}$, defined at the input level; it is the mean square error between the input and reconstructed images. Second is the sum of the reconstruction losses, $\mathcal{L}_{int}$, at the intermediate layers; they are the mean square errors between the maps of the corresponding encoders and decoders. Here two more comparison algorithms are implemented. \texttt{UNet-C-multi} linearly combines the supervised loss with the reconstruction loss at the input level without considering those defined at the intermediate layers whereas \texttt{UNet-C-multi-int} also considers the latter losses. Here two variants are implemented since it becomes harder to select the right contribution of each loss in the joint loss function as the number of losses increases. These variants are to better understand this phenomenon. 

\begin{figure}[t!]
\centering
\small{
\begin{tabular}{@{~}c@{~~~~~~~~~}c@{~}}
\includegraphics[width=0.30 \columnwidth]{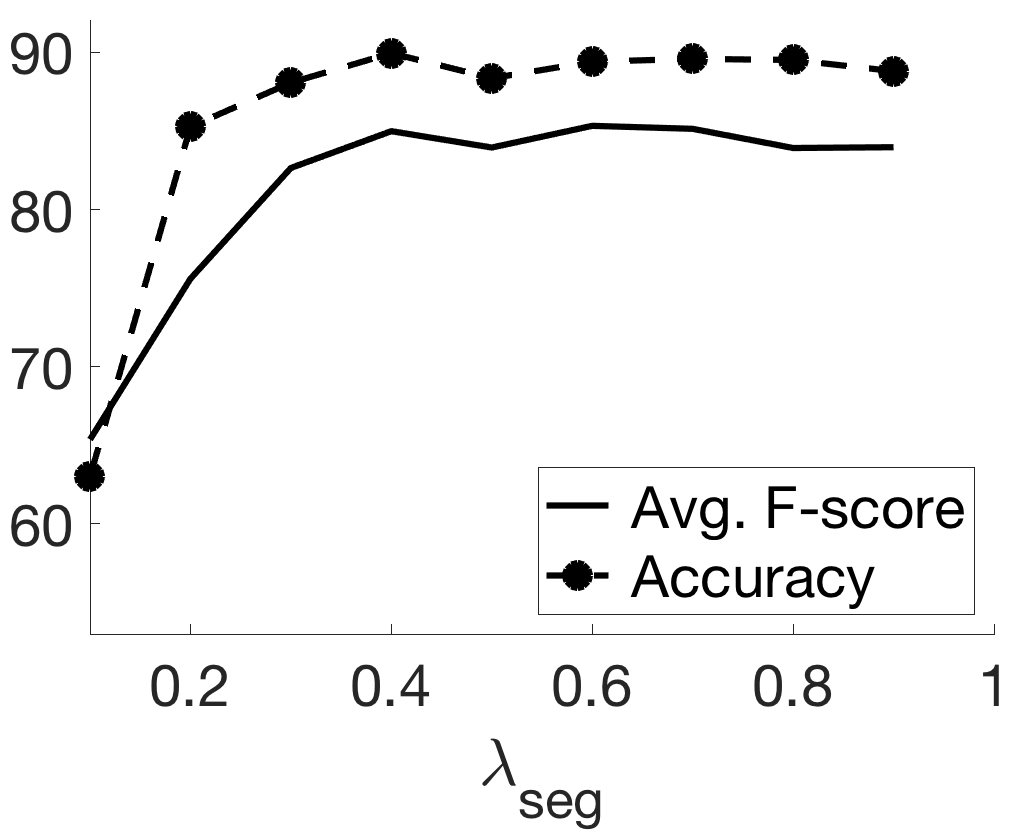} &
\includegraphics[width=0.30 \columnwidth]{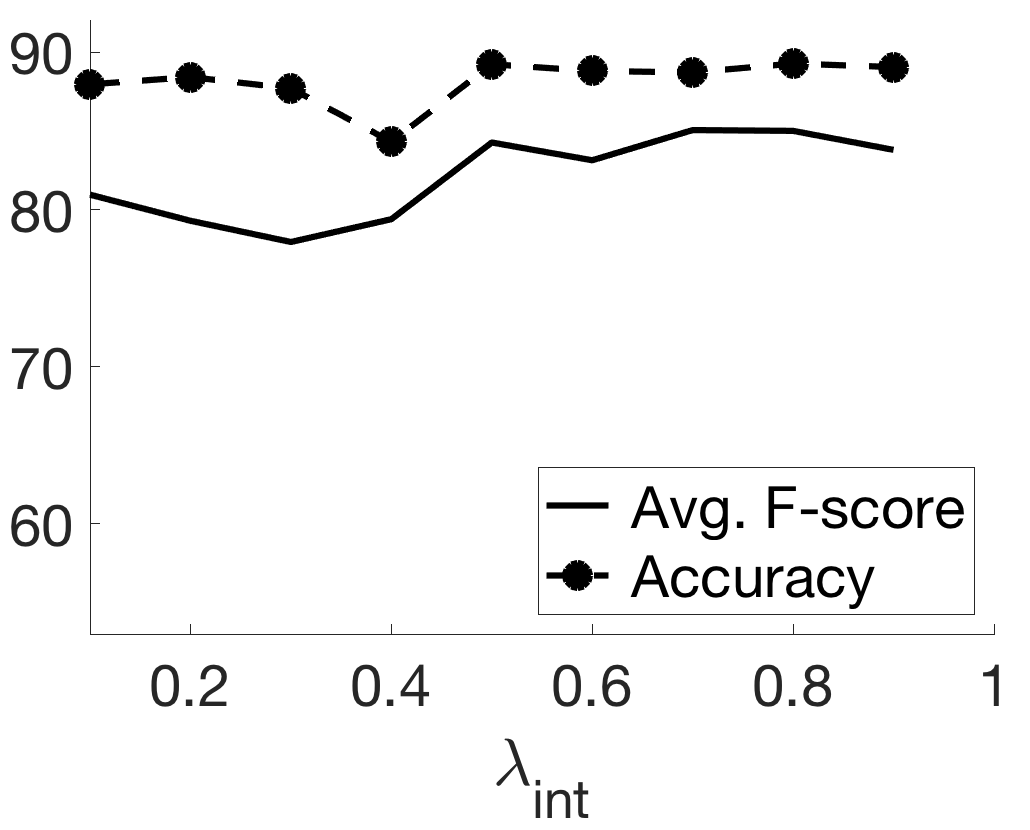} \\
(a) & (b)
\end{tabular}
}
\caption{For the in-house colon dataset, accuracy and average F-scores of (a) \texttt{UNet-C-multi} as a function of $\lambda_{seg}$, and (b) \texttt{UNet-C-multi-int} as a function of $\lambda_{int}$. In (b), $\lambda_{seg} = 0.6$, which gives the best average F-score for \texttt{UNet-C-multi}.}
\label{fig:mt_params}
\end{figure}

\texttt{UNet-C-multi} defines its joint loss function as
\begin{equation}
\mathcal{L}_{model} = \lambda_{seg}~\mathcal{L}_{seg} + \lambda_{rec}~\mathcal{L}_{rec}
\end{equation}
where $\lambda_{seg}$ and $\lambda_{rec}$ are the coefficients of the supervised and unsupervised losses, respectively. Here to find a good combination of these coefficients, we set $\lambda_{rec} = (1 - \lambda_{seg})$ and perform the grid search on the test set/fold images. In Fig.~\ref{fig:mt_params}(a), the metrics are plotted as a function of $\lambda_{seg}$ for the in-house colon dataset. When $\lambda_{seg}$ is too small, the performance of the segmentation task decreases dramatically. On the contrary, when it is close to 1, the image reconstruction task cannot help improve the results. The plots show similar characteristics for the other datasets. This grid search selects $\lambda_{seg} = 0.6$, which gives the best average F-score for the in-house colon dataset. Using the same approach, $\lambda_{seg} = 0.4$ is selected for the other two datasets. Table~\ref{table:results} and Figs.~\ref{fig:results-in-house} and~\ref{fig:results-others} present the results for these $\lambda_{seg}$ values. These results show that a multi-task network, which regularizes its training by simultaneously minimizing the supervised and unsupervised losses, improves the results of the single-stage networks. On the other hand, iMEMS leads to better results. The reason might be the following: First, iMEMS unites the supervised and unsupervised tasks into a single one and trains its network by minimizing the loss defined on the united task. This united task provides a very natural way of loss definition, eliminating the necessity of defining a joint loss function with right contributions of the supervised and unsupervised losses. This may provide more effective regularization for employing unsupervised learning in network training.  Second, iMEMS learns this united task by benefiting from the well-known synthesizing ability of cGANs. Thanks to using a cGAN, iMEMS produces realistic outputs that better comply with spatial contiguity.

\texttt{UNet-C-multi-int} defines a similar loss function, but this time, also considering the sum of the reconstruction losses, $\mathcal{L}_{int}$, at the intermediate layers. It defines the following joint loss function, which is also used in~\cite{zhao15,zhang16} to regularize their network training.
\begin{equation}
\mathcal{L}_{model} = \lambda_{seg}~\mathcal{L}_{seg} + \lambda_{rec}~\mathcal{L}_{rec} + \lambda_{int}~\mathcal{L}_{int}
\end{equation}
As aforementioned, as their number increases, it becomes harder to adjust the coefficients relative to each other. In our experiments, we use the best configuration of $\lambda_{seg} = 0.6$ and $\lambda_{rec} = 0.4$ selected by \texttt{UNet-C-multi} for the in-house colon dataset and $\lambda_{seg} = 0.4$ and $\lambda_{rec} = 0.6$ for the other datasets, and determine the coefficient $\lambda_{int}$ also by the grid search. This grid search gives the best average F-score when $\lambda_{int}$ is 0.8, 0.7, and 0.3, for the in-house colon, epithelium, and tubule datasets, respectively. For the in-house colon dataset, the metrics are plotted as a function of $\lambda_{int}$ in Fig.~\ref{fig:mt_params}(b). The test set/fold results for these $\lambda_{int}$ values are provided in Table~\ref{table:results}. Here it is observed that the inclusion of the intermediate layer losses does not help further improve the results. The reason might be the following: The linear function, which is used by \texttt{UNet-C-multi-int} as well as by the previous studies~\cite{zhao15,zhang16}, may not be the best way to combine these losses and/or it may require a more thorough coefficient search. On the contrary, the iMEMS method requires neither such an explicit joint loss function definition nor such a coefficient search since its proposed united task intrinsically combines these losses.

\section{Conclusion}

This paper proposed the iMEMS method that employs unsupervised learning to regularize the training of a fully convolutional network for a supervised task. This method proposes to define a new embedding to unite the main supervised task of semantic segmentation and an auxiliary unsupervised task of image reconstruction into a single task and to learn this united task by a conditional generative adversarial network. Since the proposed embedding corresponds to a segmentation map that preserves a reconstructive ability, the united task of its learning enforces the network to jointly learn image features and context features. This joint learning lends itself to more effective regularization, leading to better segmentation results. Additionally, this united task provides an intrinsic way of combining the segmentation and image reconstruction losses. Thus, it attends to the difficulty of defining an effective joint loss function to combine the separately defined segmentation and image reconstruction losses in a balanced way. We tested this method for semantic tissue segmentation on three datasets of histopathological images. Our experiments revealed that it leads to more accurate results compared to its counterparts. 

The proposed method is to segment a heterogeneous tissue image into its homogeneous regions. Thus, it can be easily applied to segmenting tissue compartments in whole slide images (WSIs), as in the case of many previous studies. To do so, a WSI can be divided into image tiles, on which the method predicts the output. Alternatively, an image window can be slid on the WSI and the estimated outputs can be averaged to obtain the final segmentation. This application can be considered as one future research direction. The focus of this paper is to segment a histopathological image into its tissue compartments. It is possible to extend this idea for the instance segmentation problem in histopathological images. This extension may require modifying the embedding such that it also covers additional supervised tasks (such as the task of predicting instance boundaries) that might be important for instance segmentation. The investigation of this possibility is considered as another future research direction. 



\section*{Acknowledgments}

This work was supported by the the Scientific and Technological Research Council of Turkey under the project number T{\"U}B\.{I}TAK 116E075. Authors thank Ms. G. N. Gunesli for her assistance in running the \texttt{UNet-C-multi} and \texttt{UNet-C-multi-int} comparison algorithms on our in-house dataset as the preliminary work of this paper. 






\end{document}